\documentclass[conference]{IEEEtran}
\IEEEoverridecommandlockouts
\usepackage{cite}
\usepackage{amsmath,amssymb,amsfonts}
\usepackage{algorithmic}
\usepackage{graphicx}
\usepackage{textcomp}
\usepackage{xcolor}

\usepackage{tabularx}
\usepackage{multirow}
\usepackage{booktabs}
\usepackage{comment}
\usepackage{footnote}
\usepackage{url}
\usepackage{hyperref}
\def\BibTeX{{\rm B\kern-.05em{\sc i\kern-.025em b}\kern-.08em
    T\kern-.1667em\lower.7ex\hbox{E}\kern-.125emX}}
\begin{document}

\title{From a Fourier-Domain Perspective\\ on Adversarial Examples to a\\ Wiener Filter Defense for Semantic Segmentation\\[.75ex] 
  {\normalfont\large 
    Nikhil~Kapoor$^{1,2}$,\hspace{0.3cm} Andreas Bär$^{2}$,\hspace{0.3cm} Serin Varghese$^{1}$,\hspace{0.3cm} Jan David Schneider$^{1}$,\hspace{0.3cm} \\[-2ex] Fabian H\"uger$^{1}$,\hspace{0.3cm}
Peter Schlicht$^{1}$,\hspace{0.3cm} Tim~Fingscheidt$^{2}$%
  }\\[-1ex]
}

\author{
    \IEEEauthorblockA{%
        \textit{$^{1}$Volkswagen Group Innovation} \\%
        Wolfsburg, Germany \\
        \tt\small \{nikhil.kapoor, john.serin.varghese,  jan.david.schneider, \\ fabian.hueger, peter.schlicht\}@volkswagen.de
    }
    \and
    \IEEEauthorblockA{%
        $^{2}$\textit{Technische Universität Braunschweig}\\
        Braunschweig, Germany\\%
        \tt\small \{n.kapooor, andreas.baer,\\ t.fingscheidt\}@tu-bs.de
    }
}

\maketitle

\begin{abstract}
Despite recent advancements, deep neural networks are not robust against adversarial perturbations. Many of the proposed adversarial defense approaches use computationally expensive training mechanisms that do not scale to complex real-world tasks such as semantic segmentation, and offer only marginal improvements. In addition, fundamental questions on the nature of adversarial perturbations and their relation to the network architecture are largely understudied. In this work, we study the adversarial problem from a \textit{frequency domain perspective}. More specifically, we analyze discrete Fourier transform (DFT) spectra of several adversarial images and report two major findings: First, there exists a \textit{strong connection between a model architecture and the nature of adversarial perturbations} that can be \textit{observed and addressed in the frequency domain.} Second, the observed \textit{frequency patterns are largely image- and attack-type independent, which is important for the practical impact of any defense making use of such patterns.} Motivated by these findings, we additionally propose an adversarial defense method based on the well-known Wiener filters that captures and suppresses adversarial frequencies in a data-driven manner. Our proposed method not only generalizes across unseen attacks but also excels five existing state-of-the-art methods across two models in a variety of attack settings. 
\end{abstract}

\begin{IEEEkeywords}
adversarial defense, robustness, semantic segmentation, wiener filtering
\end{IEEEkeywords}

\section{INTRODUCTION}
\begin{figure}[t!]
    \centering
    \includegraphics[width=\linewidth]{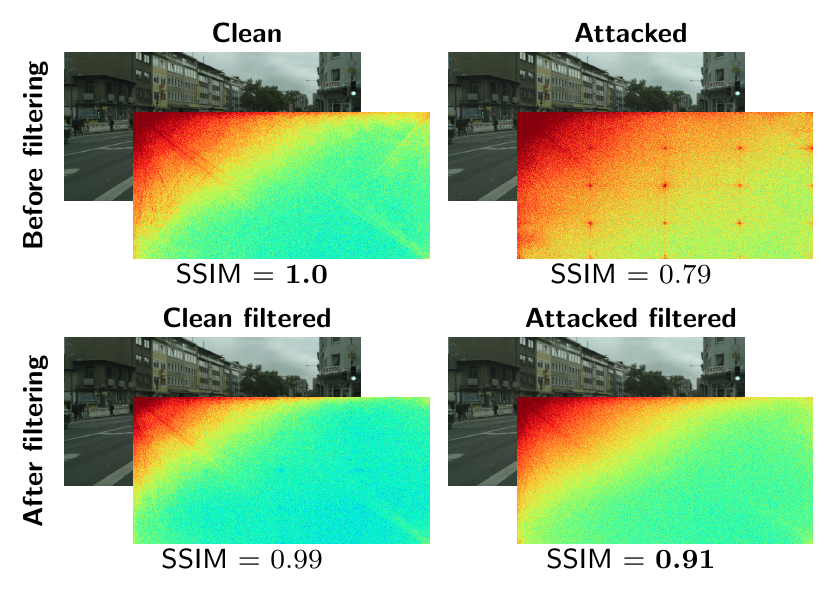}
    \caption{\textbf{Visualization of the effect of Wiener filtering} on clean data (left) and perturbed/attacked data (right). Background: image data; foreground: respective unshifted DFT spectrum. \textit{Strong grid-shaped artifacts} are seen in the DFT spectrum of attacked images on ICNet~\cite{Zhao2018a} (top right) which are suppressed after Wiener filtering (bottom right). Wiener filtering substantially improves the SSIM on attacked images (right side) and retains a reasonably high SSIM on clean images (left side).}
    \label{fig:frequency_spectra_example}
\end{figure}{}
\noindent Szegedy \textit{et al.}~\cite{Szegedy2014} pointed out that state-of-the-art machine learning models~\cite{Krizhevsky2012, Chen2017, Loehdefink2019} can be easily fooled by adding carefully crafted adversarial perturbations to the input. Although these so-called adversarial examples are often imperceptible, they cause the model to make erroneous predictions with a high confidence~\cite{Nguyen2015, Baer2021}. Since then, significant research efforts have been made in trying to get a deeper understanding into the existence of such adversarial attacks~\cite{Goodfellow2015, Chakraborty2018}, and in building reliable defense methods against such attacks~\cite{Baer2019, Baer2020, Klingner2020}. Unfortunately, this problem is still far from solved. 

Most existing defense methods are either computationally too expensive~\cite{Goodfellow2015, Madry2018}, provide a false sense of security~\cite{Papernot2016, Athalye2018}, or offer only marginal improvements in a controlled setting~\cite{Guo2018, Shaham2018}. 

To obtain certifiable, effective, and low-complex defense methods, we need to shift our focus towards answering fundamental questions to improve our understanding of this adversarial robustness problem. For example,~\cite{Assion2019} made significant advancements in understanding adversarial attacks by proposing a structured taxonomy, specifying key components of an attack. In addition, \cite{Yin2019} and \cite{Wang2020_1, Wang2020_2} analyzed the robustness problem from a frequency-domain perspective to get a better understanding of trade-offs between robustness and performance and indicated that the lack of robustness is most likely correlated to a network's sensitivity to high-frequency information in the input. Inspired from their work, we investigate adversarial attacks deeper in a frequency-domain formulation. Unlike existing analyses, we do this for a complex task, namely, semantic segmentation. 

Our initial investigation revealed that \textit{adversarial images tend to show strong artifacts in the frequency domain that are otherwise absent in their clean counterparts} (see Fig.~\ref{fig:frequency_spectra_example}). Further analysis with multiple attacks and models revealed two key findings: First, \textit{the observed adversarial patterns in the frequency domain are largely image-and attack-type-independent.} Second, \textit{there exists a strong connection between a model architecture and the nature of adversarial perturbations which can be observed and addressed in the frequency domain.} 

Motivated by these findings, we propose a new adversarial defense method using the widely known Wiener filters~\cite{Scalart1996, Kazubek2003} from image and speech denoising domain. We argue that Wiener filters are suitable for adversarial image denoising, as these filters tend to remove adversarial noise from attacked images in the frequency domain, while at the same time, leaving clean images largely unaffected (see Fig.~\ref{fig:frequency_spectra_example}). In comparison to existing denoising methods~\cite{Buades2005, Dziugaite2016, Xu2017, Aydemir2018}, \textit{Wiener filtering tends to outperform most of the methods in a variety of different attack settings}.
In summary, our results indicate that Wiener filtering tends to have many favorable properties from an adversarial defense perspective. More specifically, \textit{Wiener filter based denoising tends to generalize across unseen attack types and works even under varying attack settings} such as attack strength or $L_{p}$ norm mismatch. We hypothesize that most of these properties are primarily because the frequency patterns that the Wiener filters try to suppress are already largely attack-type and attack-setting agnostic. 

\section{RELATED WORK}
\label{section:Related_Work}
In this section, we discuss the related work on adversarial defense methods with a focus on denoising methods.

\textbf{Adversarial Defenses:} Most adversarial defenses fall into three categories: The first one is called \textit{adversarial training}~\cite{Goodfellow2015}, which includes expensive retraining of the network on adversarial examples. Although successful, this does not easily scale to complex tasks and huge data sets~\cite{Kurakin2017a}. The second category is called \textit{gradient masking}~\cite{Papernot2017}, where the idea is to hide gradients from a potential hacker~\cite{Madry2018, Cisse2017, Kannan2018}. However, \cite{Athalye2018} showed that none of these approaches are truly effective, as the hacker can compute gradients using different models and still be successful. 

\textbf{Denoising-based Defenses:} The third category is called \textit{input transformations}~\cite{Guo2018}, where the main idea is to remove adversarial perturbations directly from the input before it is fed into the model. A few well-known methods of this type include image compression methods, such as JPEG~\cite{Dziugaite2016} and JPEG2000~\cite{Aydemir2018} compression, and various smoothing methods that reduce the feature space of the input, such as non-local (NL) means~\cite{Buades2005}, bit-depth reduction~\cite{Xu2017} and median smoothing~\cite{Xu2017}. Other interesting approaches include non-differentiable smoothing methods, such as quilting and total variance minimization~\cite{Guo2018}, image cropping~\cite{Graese2016}, different forms of randomness~\cite{Xie2018, Cohen2019, Raff2019}, basis function transforms~\cite{Shaham2018}, and more advanced methods, such as high-level guided denoiser~\cite{Liao2018} and BlurNet~\cite{Raju2019}. 

In this work, we propose to use Wiener filters as an \textit{input transformation} defense method that performs denoising in the frequency domain rather than in the spatial domain. For the state-of-the-art comparison, we limit our investigations to \textit{spatial-domain based denoising methods that do not need additional retraining}. We choose five such methods that are often cited in the defense literature~\cite{Xu2017, Shaham2018, Das2018, Tang2019} including two image compression methods, namely JPEG~\cite{Dziugaite2016} and JPEG2000~\cite{Aydemir2018} compression and three feature squeezing smoothing methods, namely median blurring~\cite{Xu2017}, non-local (NL) means~\cite{Buades2005} smoothing, and bit-depth reduction~\cite{Xu2017}.
\vspace{2mm}

\section{WIENER FILTER AS ADVERSARIAL DEFENSE}
\label{section:Method}
Wiener filters are widely used as denoising filters in traditional signal processing \cite{Smith2003, Petrou2010}. In this section, we introduce an adaptation of the Wiener filter as an adversarial defense which is achieved by estimating the power spectrum of the underlying adversarial perturbations in the frequency domain. But first, we introduce some mathematical notation.

\begin{figure}[t!]
    \centering
    \includegraphics[]{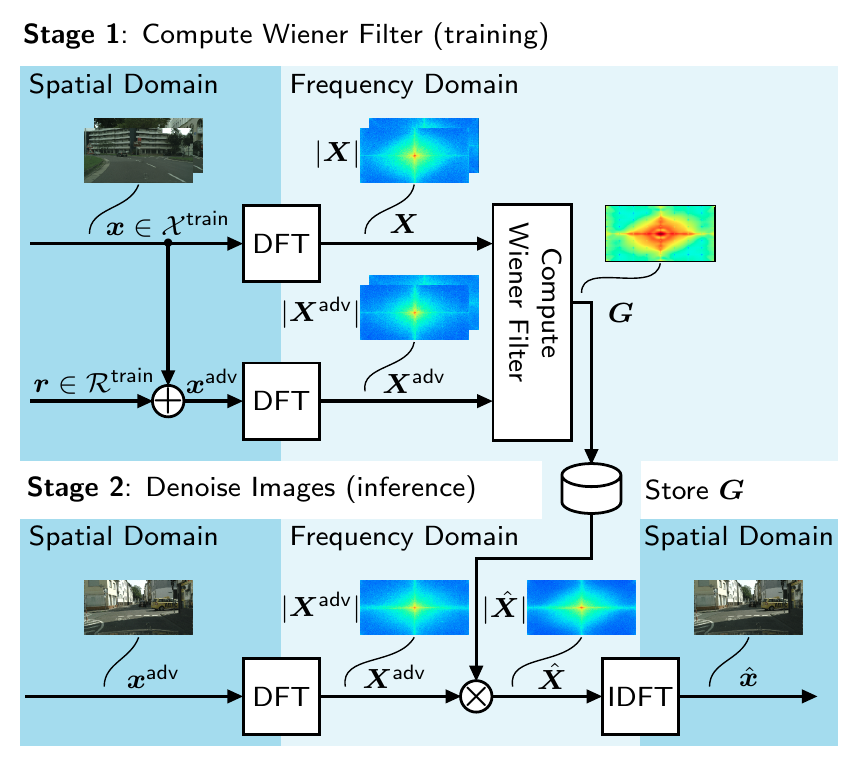}
    \caption{\textbf{Method Overview.} In\textbf{ stage 1}, a Wiener filter is computed in the frequency domain using a paired set of images $\boldsymbol{x} \in \mathcal{X}^{\text{train}}$ taken from the training set $\mathcal{X}^{\text{train}}$ with their respective adversarial perturbations $\boldsymbol{r}$ or rather their adversarial images $\boldsymbol{x}^{\text{adv}}$. In \textbf{stage 2}, during inference, the Wiener filter $\boldsymbol{G}$ is applied as an input preprocessing step such that it suppresses adversarial frequencies (if present) in the frequency domain leading to a denoised image $\hat{\boldsymbol{x}}$ in the spatial domain after an inverse DFT.}
    \label{fig:method_overview}
\end{figure}{}

\subsection{Mathematical Notations}
Let $\boldsymbol{x}\in\mathbb{G}^{H\times W\times C}$ be an image in the spatial domain with height $H$, width $W$, $C=3$ color channels, set of integer gray values $\mathbb{G}$, and $\boldsymbol{x}\in\mathcal{X}$, where $\mathcal{X}$ represents a dataset. In the frequency domain, we define $\boldsymbol{X}=(X_{k,\ell,m}) \in\mathbb{C}^{H\times W \times C}$, with $X_{k, \ell, m}$ being the element of $\boldsymbol{X}$ at frequency coordinates $k,\ell,m$ along height, width, and color channels, respectively, and $\mathbb{C}$ denoting the set of complex-valued numbers. Further, $\boldsymbol{X}$ is obtained by computing the 3D discrete Fourier transform (DFT) following \cite{Gonzalez2002} as
\begin{equation}
    X_{k,\ell,m} = \sum_{h=0}^{H-1} \sum_{w=0}^{W-1} \sum_{c=0}^{C-1} x_{h, w, c}e^{-j2\pi(\frac{k}{H}h + \frac{\ell}{W}w + \frac{m}{C}c)},
    \label{eq:DFT}
\end{equation}
where, similarly to the frequency domain, $x_{h,w,c}$ are the elements of $\boldsymbol{x}$ at spatial coordinates $h,w,c$ in height, width and color channels, respectively. For the sake of simplicity, we denote the 3D DFT with the notation $\boldsymbol{X}~=~ \mathcal{F}(\boldsymbol{x})$. Similarly, the 3D inverse DFT is denoted as $\boldsymbol{x} = \mathcal{F}^{-1}(\boldsymbol{X})$. For the sake of completeness, the 3D inverse DFT following \cite{Gonzalez2002} is defined as 
\begin{equation}
    x_{h,w,c} = \frac{1}{HWC}\sum_{k=0}^{H-1} \sum_{\ell=0}^{W-1} \sum_{m=0}^{C-1} X_{k, \ell, m}e^{j2\pi(\frac{h}{H}k + \frac{w}{W}\ell + \frac{c}{C}m)}.
    \label{eq:IDFT}
\end{equation}

The image $\boldsymbol{x}$ is fed as an input to a semantic segmentation neural network $\mathfrak{F}(\cdot)$ having network parameters $\boldsymbol{\theta}$ and output $\boldsymbol{y} = \mathfrak{F}(\boldsymbol{x}, \boldsymbol{\theta})\in[0,1]^{H\times W\times |\mathcal{S}|}$, with the set of classes $\mathcal{S}$, and $|\mathcal{S}|$ being the number of classes. Each element of $\boldsymbol{y}$ is considered to be a posterior probability $y_{i,s}=P(s | i, \boldsymbol{x})$ for the class $s\in\mathcal{S}$ at pixel index $i\in\mathcal{I}=\lbrace 1, 2, ...,H\cdot~W \rbrace$. 

An adversarial attack comprises the computation of an adversarial perturbation $\boldsymbol{r}\in\mathbb{R}^{H\times W \times C}$, such that $||\boldsymbol{r}||_{p} \leq \epsilon$,
with $\epsilon \in \mathbb{R}^{+}$ defined as an imperceptibility constraint based on the $L_{p}$ norm, denoted by $||\cdot||_{p}$. Further, an adversarial example is defined as $\boldsymbol{x}^{\text{adv}} = \boldsymbol{x} + \boldsymbol{r}.$

These adversarial perturbations $\boldsymbol{r}$ can be computed for multiple images individually, creating a set $\mathcal{R}$ of adversarial perturbations with $\boldsymbol{r}\in\mathcal{R}$.

\subsection{Defense Objective}
In the absence of an adversarial attack detector, we aim at designing a denoising filter that removes an adversarial perturbation $\boldsymbol{r}$ from an adversarial image $\boldsymbol{x}^{\text{adv}}$, and at the same time to the largest extent preserves the clean image $\boldsymbol{x}$. Mathematically speaking, if our denoising function is denoted by $g(\cdot):\mathbb{G}^{H\times W\times C} \to \mathbb{G}^{H\times W\times C}$, then ideally, the objective of denoising can be defined as
\begin{equation}
    \forall \boldsymbol{x}, \boldsymbol{x}^{\text{adv}}: g(\mathbf{x}) = \mathbf{x} \wedge g(\mathbf{x}^{\text{adv}}) = \mathbf{x}.
\end{equation}

A denoising function that satisfies the abovementioned properties has direct applications as an input preprocessing step during inference. Since the function $g(\cdot)$ will not alter a clean image $\boldsymbol{x}$ significantly, in contrast to an adversarial image $\boldsymbol{x}^{\text{adv}}$, we do not need to detect whether the input image is adversarial or not. In addition, we also do not want the original baseline performance on clean filtered images to suffer too much.

\subsection{Wiener Filter}
\noindent Wiener filters~\cite{Wiener1949} are used primarily as a denoising technique in traditional signal processing both for vision~\cite{Kazubek2003, Kethwas2015} and speech~\cite{Scalart1996, Meyer2020}. These filters typically operate in the frequency domain and assume that the spectral properties of the original images as well as the noise are known or can be estimated. 

In the context of adversarial attacks, we consider the clean images $\boldsymbol{x}$ as the original signal, which is corrupted by additive noise (also called adversarial perturbations $\boldsymbol{r}$) giving a degraded output image $\boldsymbol{x}^{\text{adv}}$. In the DFT domain, an adversarial example can be written as $\mathcal{F}(\boldsymbol{x}^{\text{adv}}) = \mathcal{F}(\boldsymbol{x}) + \mathcal{F}(\boldsymbol{r})$,
where $\mathcal{F}(\cdot)$ is the 3D DFT computed from (\ref{eq:DFT}). 
A Wiener filter in the DFT domain typically results in a transfer function $\boldsymbol{G} \in [0,1]^{H \times W \times C}$, such that the mean squared error 
between the estimated image in the DFT domain $\hat{\boldsymbol{X}}$ and the original clean image $\boldsymbol{X}$ is minimized~\cite{Gonzalez2002}. The following assumptions are made in the DFT domain: The perturbation $\boldsymbol{R} = \mathcal{F}(\boldsymbol{r})$ and the image $\boldsymbol{X}$ are uncorrelated (can be assumed due to their transferability~\cite{Demontis2019}), zero mean, and the pixel values in the estimate $\hat{\boldsymbol{X}}$ are a linear function of the pixel values in the degraded image $\boldsymbol{X}^{\text{adv}}$. Based on these assumptions, the Wiener filter transfer function $\boldsymbol{G}$ in the DFT domain is computed by
\begin{equation}
    G_{k,\ell,m} = \frac{|X_{k,\ell,m}|^2}{|X_{k,\ell,m}|^2 + |R_{k,\ell,m}|^2},
    \label{eq:wiener_filter_h}
\end{equation}
    where $G_{k,\ell,m}$ refers to the elements of $\boldsymbol{G}$ at frequency coordinates $k,\ell,m$ respectively, and $|X_{k,\ell,m}|^2$,  $|R_{k,\ell,m}|^2$ refer to the power spectrum of the clean image and the additive noise, respectively. Given this filter, one could compute the denoised image in the DFT domain by 
\begin{equation}
        \hat{\boldsymbol{X}} = \boldsymbol{G} \odot \boldsymbol{X}^{\text{adv}},
        \label{eq:wiener_filter_multiplication}
\end{equation}
where $\odot$ represents an element-wise multiplication. Finally, the denoised image in the spatial domain could be recovered by simply computing the inverse DFT (IDFT) as in (\ref{eq:IDFT}):
\begin{equation}
\hat{\boldsymbol{x}} = \mathcal{F}^{-1}(\hat{\boldsymbol{X}}).
\label{eq:IDFT_F}
\end{equation}

The challenge in real-world denoising applications is now to estimate the clean signal power spectrum and the noise power spectrum, or, alternatively, the signal-to-noise ratio $SNR_{k,\ell,m} = |X_{k,\ell,m}|^2/|R_{k,\ell,m}|^2,
    \label{eq:snr}$ allowing (\ref{eq:wiener_filter_h}) to be rewritten as 
\begin{equation}
    G_{k,\ell,m} = \frac{SNR_{k,\ell,m}}{1+SNR_{k,\ell,m}}.
    \label{eq:wiener_filter_h_snr}
\end{equation}

For our application, we can state that the filter defined in (\ref{eq:wiener_filter_h}) optimally works for a single known image and known noise. In an experimental setting, where $X_{k,\ell,m}$ and $R_{k,\ell,m}$ are available separately, one can compute (\ref{eq:wiener_filter_h}) and apply (\ref{eq:wiener_filter_multiplication}) and (\ref{eq:IDFT}) to achieve an \textit{upper performance limit} later on. For a real-world input transformation, however, we adapt this to unknown images in inference for an attack type $a$, by simply taking the arithmetic mean over multiple images from a training set $\boldsymbol{x} \in \mathcal{X}^{\mathrm{train}}=\mathcal{X}$ as follows
\begin{equation}
   G^{(a)}_{k,\ell,m} = \frac{1}{|\mathcal{X}|}\sum_{\boldsymbol{x}\in\mathcal{X}}{} \frac{SNR_{k,\ell,m}^{(a)}}{1 + SNR_{k,\ell,m}^{(a)}},  
    \label{eq:wiener_filter_single_attack}
\end{equation}
with the superscript $a$ referring to the respective attack type used for creating the perturbation $R_{k,\ell,m}^{(a)}$ to obtain $SNR_{k,\ell,m}^{(a)}=|X_{k,\ell,m}|^{2}\big{/}|R_{k,\ell,m}^{(a)}|^{2}$, and then to obtain the single-attack Wiener filter $G_{k,\ell,m}^{(a)}$, or $\boldsymbol{G}^{(a)}$. This averaging approach is indeed novel and ambitious for combating adversarial perturbations, as it requires a similar SNR for different images at a particular DFT bin for a certain attack type $a$, as we will show in Section~\ref{section:Experimental_Results}. Similarly, we can even combine filters from multiple attack types $a \in \mathcal{A}$ from a set of attacks $\mathcal{A}$ to obtain a single Wiener filter $\boldsymbol{G}^{(\mathcal{A})}$ with coefficients
\begin{equation}
    G^{\mathcal{(A)}}_{k,\ell,m} = \frac{1}{|\mathcal{A}|}\sum_{a \in \mathcal{A}}^{}{G^{(a)}_{k,\ell,m}}.
    \label{eq:wiener_filter_combined}
\end{equation}

Note that this Wiener filter computation is novel as well and even more ambitious, as the aforementioned assumptions leading to (\ref{eq:wiener_filter_single_attack}) are augmented by the assumption that the similar SNR even holds for different attack types. In other words, we expect $SNR_{k,\ell,m}^{(a)}$ for a certain network topology to be similar for various attack types $a\in\mathcal{A}$.

\section{EXPERIMENTAL SETUP}
\label{section:Experimental_Setup}
In this section, we describe the most important details of our experimental setup. More details can be found in the supplementary material.

\subsection{Employed Datasets and Models}
We perform our experiments on a well-known publically available data set for semantic segmentation, namely Cityscapes~\cite{Cordts2016}. This data set consists of $5,\!000$ RGB images with resolution $2048\times1024$ of urban street scenes from different cities across Germany. The data split comprises $2,\!975$ training images, $500$ validation images, and $1,\!525$ test images. We report our results on the clean and adversarially perturbed Cityscapes validation set using the (mean) intersection over union (IoU).

Regarding models, we focus on two well-known semantic segmentation models, namely FCN~\cite{Long2015} and ICNet~\cite{Zhao2018a}. The FCN employs fully convolutional layers and is considered as one of the pioneer architectures for semantic segmentation inspiring many future works~\cite{Zhang2018a, Chen2015, Yu2016, Chen2018}. For our work, we trained the FCN-VGG16 variant from scratch on the Cityscapes dataset for $120$ epochs with a learning rate of $0.0001$ and batch size $4$ using two \texttt{Nvidia GTX 1080Ti} GPUs and achieved $48.48\%$ mIoU on the validation set. 

On the other hand, the ICNet model~\cite{Zhao2018a} employs a multi-stream architecture which is capable of real-time inference at $30$ fps. The architecture employs cascade image inputs (i.e. low, medium and high resolution) with separate streams for each resolution combined with a feature fusion unit that outputs a full resolution segmentation map. For our work, we use an openly available pre-trained Tensorflow implementation~\cite{HZYang} of the ICNet that achieves $67.25\%$ mIoU on the validation set.
\begin{table}[t!]
\centering
\caption{\textbf{Attack parameters} used in this work.}
\label{tab:attack_parameters}
\resizebox{\linewidth}{!}{
\renewcommand{\arraystretch}{1.5}
\begin{tabular}{cccccc} 
\toprule
\multirow{2}{*}{\textbf{Attack}} & \multirow{2}{*}{\textbf{Target}} & \multirow{2}{*}{\textbf{Parameter}} & \multicolumn{2}{c}{\textbf{Epsilon $\epsilon$}} & \textbf{\# of} \\
& & & $L_{\infty}$ & $L_{2}$ & \textbf{Iterations}\\ \hline 

mFGSM & Car & \multirow{2}{*}{$\mu=10$} & 5, 10 & 5000 & 20 \\
\cite{Dong2018} & Pedestrian &  & 5, 10 & - & 20 \\\hline
\multirow{2}{*}{Metzen~\cite{Metzen2017}} & Car & - & 5, 10, 40 & - & 20 \\
& Pedestrian & - & 5, 10, 40 & - & 20 \\\hline
IM~\cite{Metzen2017} & - & - & 5, 10, 40 & - & 20 \\\hline
\multirow{2}{*}{Mopuri~\cite{Mopuri2018}} & \multirow{2}{*}{-} & {$\mu=1$} & \multirow{2}{*}{5, 10} & - & \multirow{2}{*}{20} \\
& & $\text{conv layer } \lambda=3$ & & \\
\bottomrule
\end{tabular}}
\end{table}
\begin{table}[t!]
\centering
\caption{\textbf{Method parameters} used in this work.}
\label{tab:method_parameters}
\resizebox{0.8\linewidth}{!}{
\renewcommand{\arraystretch}{1.5}
\begin{tabular}{ll} 
\toprule
\textbf{Method} & \textbf{Parameters} \\
\midrule
JPEG ~\cite{Dziugaite2016} & quality = 90 \\
JPEG2000~\cite{Aydemir2018} & quality = 400 \\
\multirow{2}{*}{NL Means (NLM)~\cite{Buades2005}} & search window = $13 \times 13$, \\
 & size = $3 \times 3$, strength = 2 \\
Median Blur (MB)~\cite{Xu2017} & filter size = $3 \times 3$ \\
Bit-Depth Reduction (BDR)~\cite{Xu2017} & \# of bits = 5 \\
\bottomrule
\end{tabular}}
\end{table}
\subsection{Adversarial Attacks}
\label{subsec:BPDA}
\begin{figure*}[t]
    \centering
    \includegraphics[width=17cm]{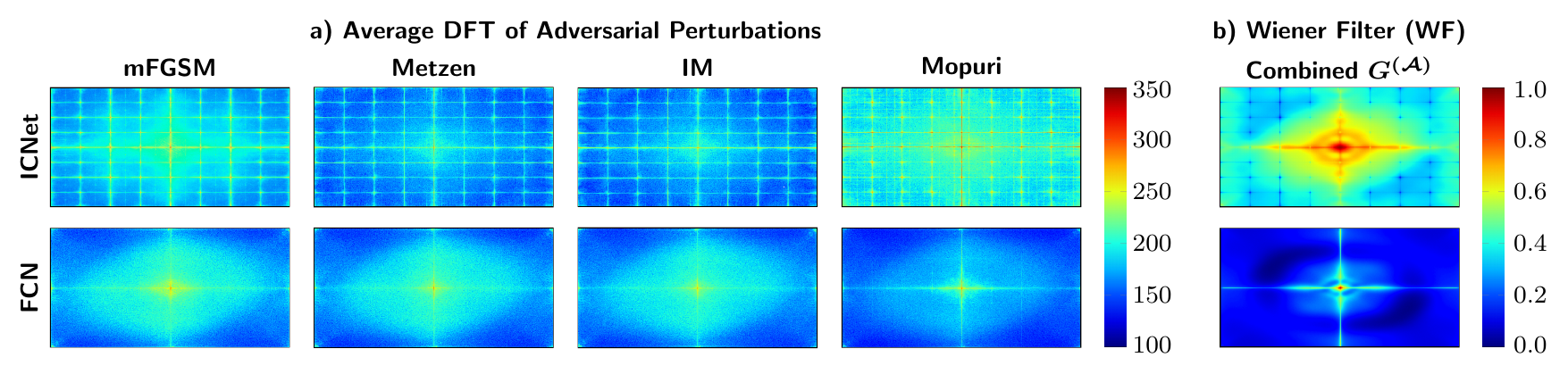}
    \caption{\textbf{Visualization of a) average amplitude spectra of adversarial perturbations} and \textbf{b) the corresponding combined Wiener filter} $G^{(\mathcal{A})}$~(\ref{eq:wiener_filter_combined}) for ICNet and FCN. For each adversarial attack, we estimate $\mathbb{E}[|\mathcal{F}(\boldsymbol{x}^{\text{adv}} - \boldsymbol{x})|]$ over the entire $2,975$ Cityscapes training set images. The DFT spectra are log-scaled such that low-frequency points are at the center of the images.}
    \label{fig:amplitude_spectrum}
\end{figure*}
\begin{figure*}[t]
    \centering
    \includegraphics[width=17cm]{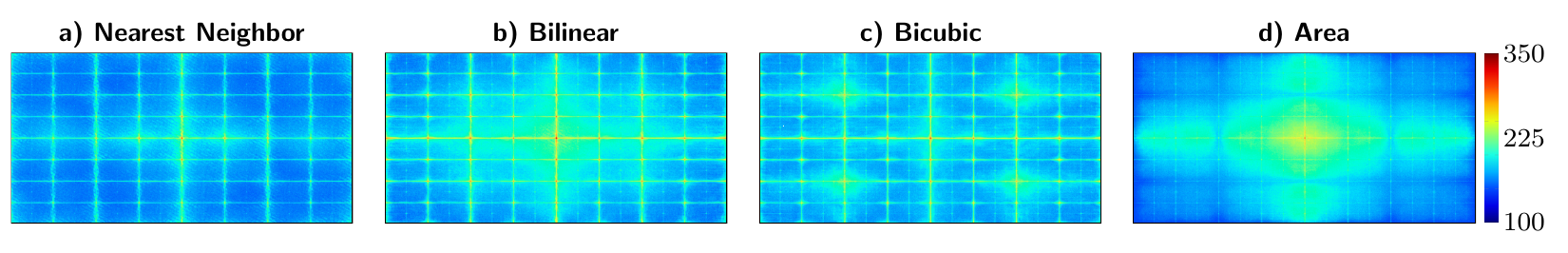}
    \caption{\textbf{Visualization of average amplitude spectra of adversarial perturbations with four different interpolation methods}, i.e. \textbf{a) nearest neighbor b) bilinear interpolation c) bicubic interpolation d) area interpolation} computed for the mFGSM attack on the \textbf{ICNet} model. We estimate $\mathbb{E}[|\mathcal{F}(\boldsymbol{x}^{\text{adv}} - \boldsymbol{x})|]$ over the entire $2,975$ Cityscapes training set images.}
    \label{fig:upsampling_2}
\end{figure*}
We employ four kinds of adversarial attacks for semantic segmentation; two targeted attacks, namely the least likely method (LLM) by Metzen~\cite{Metzen2017} and momentum FGSM (mFGSM)~\cite{Dong2018}; one untargeted attack, namely the data-free method by Mopuri et al.~\cite{Mopuri2018}; and one confusion-based attack, namely iterative mirror (IM)~\cite{Metzen2017}. These attacks are adopted from the \texttt{AidKit}\footnote{\url{https://aidkit.ai/}} attack framework~\cite{Assion2019}. The attack parameters used in this work are shown in Tab.~\ref{tab:attack_parameters} (see supplementary material for further details on all attacks).

\textbf{Backward Pass Differential Approximation (BPDA):}
The authors of \cite{Athalye2018} proposed the BPDA algorithm to generate adversarial examples against pre-processing based defenses that are either hard or impossible to differentiate. Following ~\cite{Shaham2018}, we test all defense methods in a \textit{pure white-box setting}, employing the mFGSM attack ($\epsilon=10$, $L_{\infty}$ norm, $T=20$ steps) either with or without BPDA. 

\subsection{Baseline Defenses}
For each baseline method, we use the best parameter settings adopted either from literature~\cite{Xu2017} or by using a grid search algorithm, as reported in Tab.~\ref{tab:method_parameters}.

\section{RESULTS AND DISCUSSION}
\label{section:Experimental_Results}
In this section, we discuss two sets of experiments. First, we evaluate adversarial examples in the frequency domain and report our findings and observations. Second, we perform an in-depth evaluation of the Wiener filters as adversarial defense methods and compare them to existing methods.
\subsection{Experiments with Fourier-Domain Analyses}
We investigate adversarial images in the frequency domain and compare it to their clean counterparts. First, in Fig.~\ref{fig:frequency_spectra_example} we visualized the amplitude spectrum of a \textit{single} sample clean image and its corresponding attacked image (using the mFGSM attack computed on the ICNet model with an $L_{\infty}$ norm and $\epsilon=10$). We observed that there exist \textit{several periodic grid-shaped artifacts} in the \textit{frequency domain of attacked images} which were absent in the clean counterparts. 

Intrigued by these artifacts, we visualized the averaged DFT spectra over all $2,\!975$ training set images for adversarial perturbations computed using the same mFGSM attack and show the results in Fig.~\ref{fig:amplitude_spectrum} a) (top left). Surprisingly, we found that \textit{every single attacked image had more or less the exact same grid-shaped artifacts in the frequency domain}. We then performed a series of carefully designed ablation experiments to investigate the source of these artifacts. 

\textbf{Effect of attack strength and iterations.} We investigate the effect of varying attack strength $\epsilon\in[5,10,15,20]$ and number of iterations $T\in[10,20,30,50]$, on the artifacts. Our results indicate that the \textit{grid patterns are largely independent of the chosen attack parameters} with only small variations in artifact severity (see supplementary material). 

\begin{figure*}[t]
    \centering
    \includegraphics[width=\textwidth]{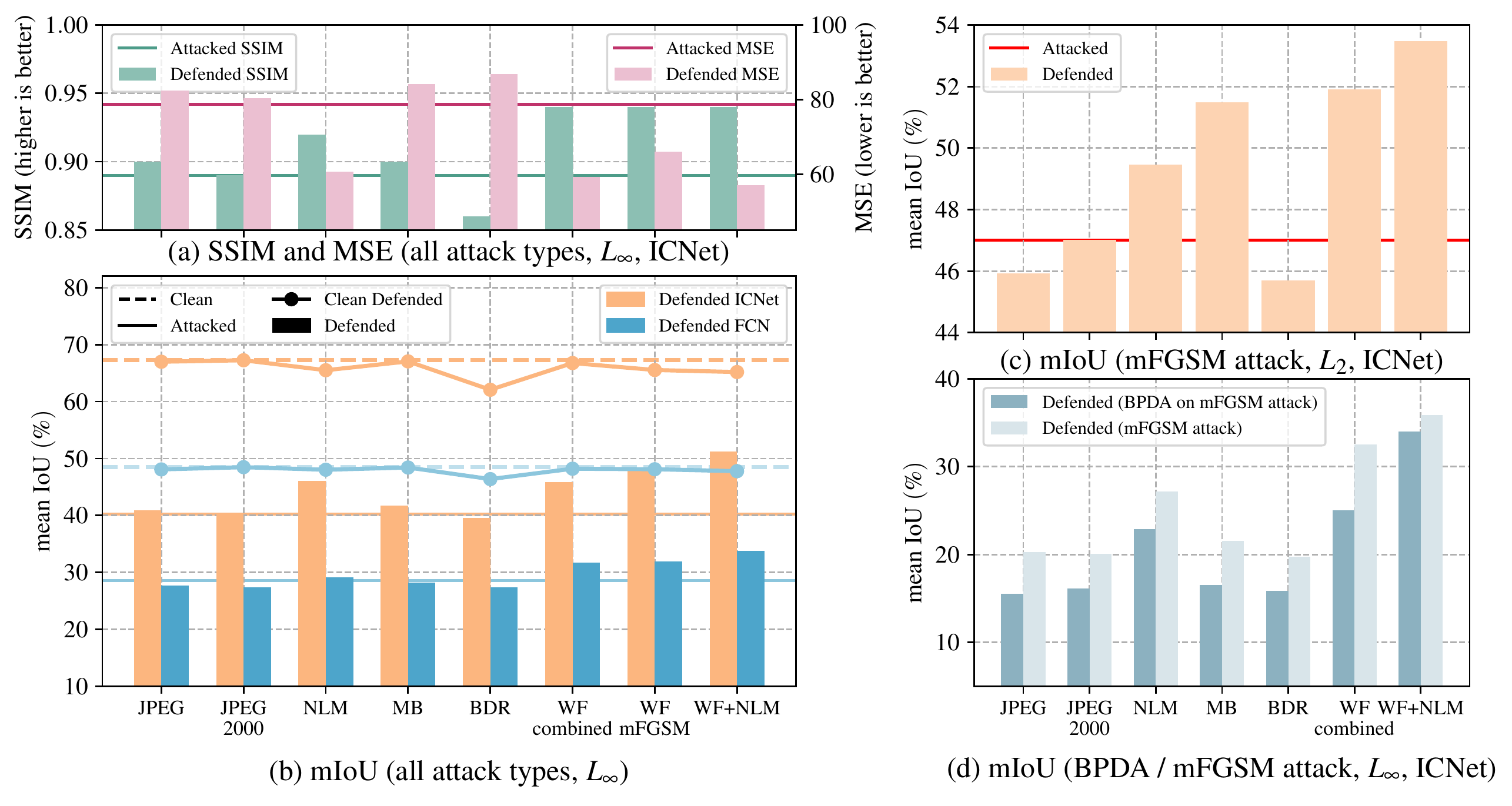}
    \caption{\textbf{Denoising Results of Wiener Filtering compared with existing state-of-the-art methods (symbols see also Table~\ref{tab:attack_parameters}}). (a) Averaged SSIM and MSE improvement across all $L_{\infty}$-based attacks on ICNet~\cite{Zhao2018a}, (b) averaged mIoU improvement all $L_{\infty}$-based attacks for ICNet and FCN, (c) mIoU improvements across $L_2$-based mFGSM attack (ICNet), (d) mIoU improvements applying BPDA on the $L_{\infty}$-based mFGSM attack (ICNet). \textbf{Note, in all plots the Wiener filter is designed for $L_{\infty}$-based attacks.}}
    \label{fig:all_results_mIoU}
\end{figure*}

\textbf{Effect of attack type.} We now study different attacks (see Tab.~\ref{tab:attack_parameters}) and visualize the corresponding averaged amplitude spectra for the same $2,\!975$ training images (results shown in Fig.~\ref{fig:amplitude_spectrum} a), top row for ICNet). Note that these attacks have fundamentally different optimization goals (i.e., targeted, untargeted, or confusion) and even different optimization techniques. However, \textit{even with these fundamental differences in the attack types, the same grid-shaped artifact was observed across all images for all attacks.} 

\textbf{Effect of network architecture.}
So far, we found that \textit{adversarial attacks tend to share inherent similarities, observed via artifacts in the frequency domain}. These artifacts are also largely \textit{image-type and attack-configuration independent}. Next, we investigate the FCN architecture and again perform similar experiments. From the results shown in Fig.~\ref{fig:amplitude_spectrum} a), bottom row, a \textit{box-shaped artifact} was observed instead. These artifacts, as before, also remained consistent across several images and attacks (see supplementary material). Hence, since different architectures lead to different artifacts, it turns out that \textit{the artifacts seem to have a strong relation to the underlying network architecture}.

\textbf{Effect of varying interpolation types.}
In the ICNet architecture, there exist six locations, where bilinear interpolation is performed to carry out either downsampling or upsampling. We investigate the effect of varying the type of interpolation used in each of the six steps, while keeping all other attack and model parameters constant. From the results shown in Fig.~\ref{fig:upsampling_2}, we make two important observations: (a) \textit{The resulting artifacts have a strong relation to different layer types of the network architecture, especially interpolation layers,} and (b) \textit{the grid-type effect is reduced as more advanced interpolation methods are used (see more details in the supplementary material).} Similar observations were reported in \cite{Frank2020} and \cite{Wang2020_3} with images generated by generative adversarial networks (GANs). Similar to GANs, an adversarial perturbation can also be assumed to be an image generated by a neural network, and hence is vulnerable to artifacts induced by different kinds of layers in the network topology.

\subsection{Why do artifacts exist in adversarial examples?}
From the literature~\cite{Lyons2014, Gazi2018, Frank2020, Wang2020_3}, there exist three possible reasons for such periodic artifacts, (a) \textit{aliasing} due to incorrect downsampling~\cite{Gazi2018}, (b) \textit{checkerboard artifacts} introduced by transposed convolutions~\cite{Odena2016}, and (c) \textit{classical interpolation} methods~\cite{Frank2020, Wang2020_3}, commonly used in multi-stream CNNs. 

An adversarial example $\boldsymbol{x}^{\text{adv}}$ consists of adding a network-generated noise pattern $\boldsymbol{r}$ on top of a clean image $\boldsymbol{x}$. This particular noise pattern, also called adversarial perturbation, is the result of a gradient-based optimization process performed on a target neural network. Typically, in iterative gradient-based attack types, several forward and backward passes are performed through different layers of a network. Hence, in each iteration, the resulting noise pattern undergoes layer-relevant changes depending on the type of layer (e.g., convolution, pooling, upsampling, deconvolution or transposed convolution, etc.), which might or might not introduce artifacts. This to some extent explains that as long as the underlying architecture remains the same, the resulting artifacts from different attack types and attack settings may also exhibit certain similarities, e.g., in the frequency domain. This might explain why we observe similar patterns from different attack types in Fig.~\ref{fig:amplitude_spectrum}. 

\subsection{Experiments with Defense Methods}
Motivated by our findings from the first set of experiments, we now investigate the benefits of using Wiener filters as a defense for semantic segmentation. 
We present results on three variants of Wiener filters, i.e., (a) the combined WF $G^{(\mathcal{A})}$~(\ref{eq:wiener_filter_combined}), also visualized in Fig.\!~\ref{fig:amplitude_spectrum} (right column), a filter averaged across all six attacks, (b) WF $G^{(a)}$~(\ref{eq:wiener_filter_h}), a single-attack filter, where the attack type $a$ corresponds to the mFGSM attack, and (c) the combined WF $G^{(\mathcal{A})}$~(\ref{eq:wiener_filter_combined}) plus NL Means (NLM) (together abbreviated as WF+NLM), where first WF is applied and then NLM on top. Detailed results on mismatched WF variants (strength $\epsilon$ / SNR mismatch) are in the supplementary material. 

\textbf{Effect on SSIM and MSE.}
In Fig.\!~\ref{fig:all_results_mIoU} (a), we investigate the defense improvements using traditional metrics such as structural similarity index metric (SSIM) and mean squared error (MSE) for attacked images $\boldsymbol{x}^{\text{adv}}$ and attacked defended image $\hat{\boldsymbol{x}}$, both w.r.t. the original clean images $\boldsymbol{x}$. Fig.\!~\ref{fig:all_results_mIoU} (a) shows averaged results across six attack types (each attack computed with an $L_{\infty}$ norm with $\epsilon=10$) for all 500 images of the Cityscapes validation set. These results indicate strong improvements with denoising using Wiener filters on SSIM and MSE, which are significantly better than other baseline methods. The WF (combined) achieves on average an impressive $5\%$ absolute SSIM improvement with a corresponding $20$ points decrease in MSE after denoising. At the same time, WF(mFGSM), although trained only on a single attack and tested across all 6 attacks (5 of which are unseen), still achieves an overall $5\%$ absolute improvement in SSIM and $13$ points decrease in MSE, indicating strong generalization properties of Wiener filters to unseen attacks. In contrast, most of the baseline methods (except NLM) barely improve the SSIM (on average $+0.87\%$ absolute), and in fact lead to a higher MSE (on average $+4.74$) after denoising. 

\textbf{Effect on mIoU ($L_{\infty}$ norm).}
In Fig.\!~\ref{fig:all_results_mIoU} (b), we investigate the improvement in mean intersection-over-union (mIoU) on all the methods and two models, namely ICNet~\cite{Zhao2018a} and FCN~\cite{Long2015} (as described in Section~\ref{section:Experimental_Setup}). Fig.\!~\ref{fig:all_results_mIoU} (b) reports averaged results over all attacks and all validation set images. For both models, attacks are computed based on the $L_{\infty}$ norm with attack strength as shown in Fig.~\ref{tab:attack_parameters}. From the results, we observe that the \textit{WF consistently offers improvements better than almost all other baseline methods on both models (on average +6\% absolute mIoU on ICNet and +3\% absolute mIoU on FCN on all attacks).} At the same time, the performance on clean images remains largely unaffected with only a $0.5\%$ absolute decrease in mIoU. However, to our surprise, WF(mFGSM) even surpasses WF(combined) with an impressive improvement of $+8\%$ absolute on ICNet and $3.36\%$ absolute on FCN, thereby indicating strong generalization towards unseen attacks. In comparison to other baselines, only NLM remains competitive, improving mIoU vs. the attacked case by an average $5.8\%$ absolute on ICNet and only $0.61\%$ absolute on FCN. All other baseline methods, in contrast to existing beliefs~\cite{Xu2017, Guo2018}, offer only marginal improvements in our complex task setting. Additionally, we investigated various combinations of denoising methods, and found that the WF can easily be combined with other methods, with the best combination being WF(combined) + NLM. \textit{Such a defense consistently excels all methods in each experiment we performed, at the cost of an only small performance drop on clean images.}

\textbf{Generalization across $L_{p}$ norms.} In a real-world setting, an attacker can vary many things to alter his attack strategy. Hence, a defense strategy that works well under such mismatched conditions is highly desirable. In Fig.\!~\ref{fig:all_results_mIoU} (c), we investigated the performance of all methods with the mFGSM attack on ICNet, computed using an $L_{2}$ norm ($\epsilon=5000$). Note that the WF used here is still computed on the $L_{\infty}$ norm instead. Even in this mismatched setting, the \textit{WF outperforms all other methods}, thereby highlighting the benefits of frequency-domain based denoising.

\textbf{Generalization across white-box attacks (BPDA).} Lastly, in Fig.\!~\ref{fig:all_results_mIoU} (d), we investigate the performance of our model in a pure \textit{white-box} setting using the BPDA attack algorithm with the mFGSM attack as described in Sec.~\ref{subsec:BPDA}. Considering fixed parameter settings ($L_{\infty}$ norm with $\epsilon=10$) for each denoising method, the attack success is reported in Fig.~\ref{fig:all_results_mIoU} (d). In absolute performance after attack, WF(combined) \textit{outperforms all other individual methods.} Finally, we observe that WF(combined)+NLM seems to be the strongest defense under all attack variations and settings that we examined.

\section{CONCLUSION}
\label{section:Conclusion}
In this paper, we performed detailed frequency-domain analysis of adversarial images across four attack types  and two network models on a complex task, namely semantic segmentation. We report two key findings: First, we observed that \textit{attacked images tend to exhibit artifacts/patterns in the frequency domain that are largely image- and attack-type independent.} Second, \textit{these artifacts originate during the attack generation process while propagating gradients across different layers of the network}, and are therefore network-dependent and particularly interpolation-type dependent. 

Motivated by these findings, we further proposed \textit{an adversarial defense method that suppresses these observed frequency patterns/artifacts using the well-known Wiener filter (WF)}. Using our method, \textit{we excel five state-of-the-art adversarial denoising methods across different $L_{p}$ norms in both gray-box and white-box attack settings}. The generalization capability of our WF can be largely attributed to the attack-agnostic properties of the underlying frequency patterns.

\section{ACKNOWLEDGEMENT}
The research leading to these results is funded by the German Federal Ministry for Economic Affairs and Energy within the project “KI Absicherung – Safe AI for Automated Driving". The authors would like to thank the consortium for the successful cooperation.

{\small
\bibliographystyle{IEEEtran}
\bibliography{references}
}

\end{document}